\documentclass{IOS-Book-Article}

\usepackage{floatrow}
\newfloatcommand{capbtabbox}{table}[][\FBwidth]
\usepackage{mathptmx}
\usepackage{soul}\setuldepth{article}
\usepackage{times}
\usepackage{soul}
\usepackage{url}
\usepackage[utf8]{inputenc}
\usepackage[small]{caption}
\usepackage{graphicx}
\usepackage{amsmath}
\usepackage{amsthm}
\usepackage{booktabs}
\usepackage[switch]{lineno}
\usepackage{tikz}

\def\hb{\hbox to 11.5 cm{}}

\begin{document}
\pagestyle{headings}
\def\thepage{}
\begin{frontmatter}              

\title{POV Learning: \\ Individual Alignment of 
Multimodal Models via Human Perception}


\author[A]{\fnms{Simon} \snm{Werner}%
\thanks{Corresponding Author: werners@uni-trier.de}},
\author[B]{\fnms{Katharina} \snm{Christ}},
\author[A]{\fnms{Laura} \snm{Bernardy}},
\author[A]{\fnms{Marion G.} \snm{Müller}}
and
\author[A]{\fnms{Achim} \snm{Rettinger}}

\address[A]{Trier University}
\address[B]{University of Innsbruck}

\begin{abstract}
Aligning machine learning systems with human preferences is mostly attempted by training with manually vetted human behavioral samples, typically explicit feedback. This is done on a population level since the context that captures the subjective Point-of-View (POV) of a concrete person in a specific situational context is not retained in the data.
We argue that alignment on an individual level can significantly boost subjective predictive performance for the individual user who interacts with the system. Naturally, perception differs for each person, thus the same situation is observed differently. Consequently, the basis for decision making and the subsequent reasoning processes and observable reactions differ for every individual.

We propose that individual perception patterns can be used to improve alignment on an individual level. We tested this by integrating perception information into machine learning systems and measured their predictive performance with respect to individual subjective assessments. For our empirical study, we collect a novel data set of multimodal stimuli and corresponding eye tracking sequences for the novel task of Perception-Guided Crossmodal Entailment and tackle it with our Perception-Guided Multimodal Transformer.
Our findings suggest that exploiting individual perception signals for machine learning of subjective human assessments provides a valuable cue for individual alignment. It not only improves the overall predictive performance from the point of view of the individual user but might also contribute to steering AI systems towards every person's individual preferences and values and ultimately improves human-AI interaction.
\end{abstract}

\end{frontmatter}

\section{Motivation}
Since the release of GPT4 \cite{openai2023gpt4} and open-source variants such as Pixtral \cite{agrawal2024pixtral12b}, multimodal transformers have received a great deal of public attention.
They show impressive capabilities on tasks like visual question answering that take image and text as input and generated text as output. However, aligning machine learning models with an individual person's Point-Of-View (POV) and preferences has not attracted much research yet.

\begin{figure}
    \begin{floatrow}
        \ffigbox{\includegraphics[scale=0.27]{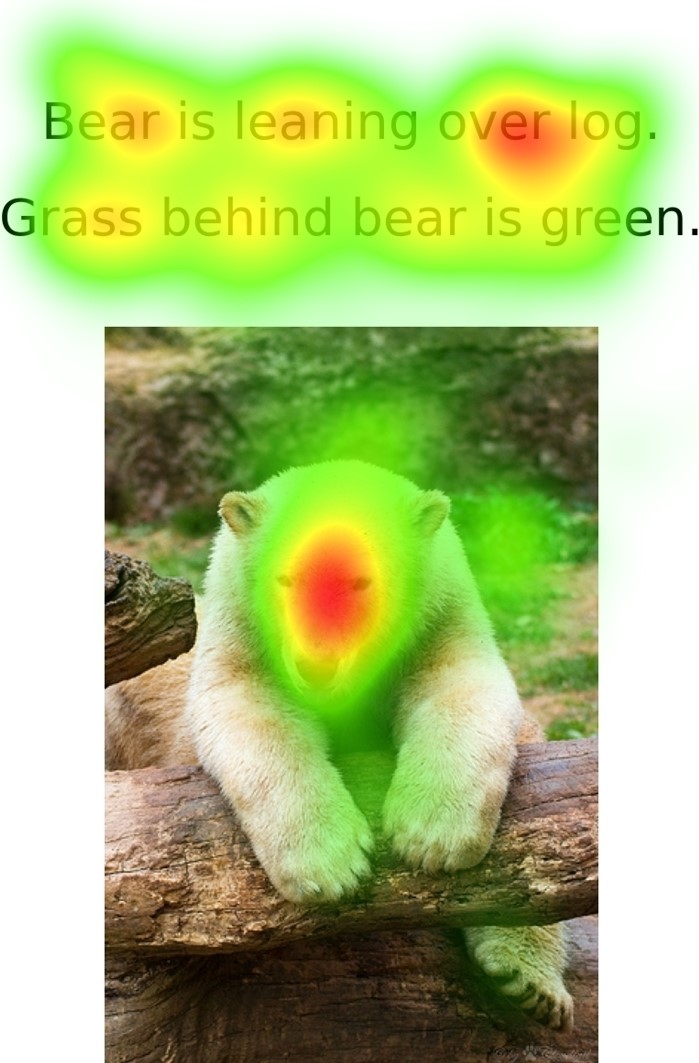}}
        {\caption{Human visual perception when given the crossmodal entailment task: \textit{Are the central objects in the image mentioned in the caption?} The red areas are indicating the focus of human attention according to eye tracking.}{}
        \label{fig:eyecatcher}}
        
        \ffigbox{\input{diagram.tex}}
        {\caption{Illustration of a \emph{POV machine learning task}: Given the \emph{stimulus} and the \emph{individual alignment signals}, a machine learning model is conditioned on the \emph{POV-specific objective} to predict the subjective inference of an individual.}%
        \label{fig:individualAlignment}}
    \end{floatrow}
\end{figure}

To improve the alignment of machine learning models according to individual expectations and preferences, we first have to capture personality cues that are potential indicators of personal preferences and values. For a stimulus consisting of text and images, the visual perception process is a potential signal that indicates an individual's evaluation of the said stimulus (see Fig.~\ref{fig:eyecatcher}).



The other required component is a machine learning objective that is an indicator of individual alignment. When accounting for tasks that measure individual behavior, a key feature is that the same stimuli might be assessed differently by different individuals. Contrary to traditional machine learning, in POV learning, this is not regarded as noise or an undesirable low inter-rater reliability. Instead, we are aiming for individual alignment, so individually differing assessments of the same stimuli is considered the key objective for optimizing the model. 
More formally, we define a \emph{POV machine learning task} illustrated in Fig.~\ref{fig:individualAlignment} as maximizing a \emph{POV-specific objective function:}

$$g^{\text{\tiny POV}}(x^{\text{\tiny POV}}) = \underset{y^{\text{\tiny POV}}}{\arg\max} \; f^\text{\tiny POV}(x^{\text{\tiny POV}},y^{\text{\tiny POV}})$$

where $y^{\text{\tiny POV}}$ is a prediction target subjective to each individual person. The given features $x^{\text{\tiny POV}}$ are a tuple of the \emph{stimulus} $x^{\text{\tiny STIM}}$ (here, text+image pairs) and the \emph{individual alignment signal} $x^{\text{\tiny ALIGN}}$ (here, eye-tracking perception traces), thus $
x^{\text{\tiny POV}} = (x^{\text{\tiny STIM}}, x^{\text{\tiny ALIGN}})$.


Based on that setup, we investigate how alignment of multimodal models on an individual level can be improved by exploiting human visual perception traces. Our contributions can be summarized as follows:
\begin{enumerate}
    \item For measuring individual alignment according to a evaluation metric $g^{\text{\tiny POV}}$, we propose the novel POV-task of Perception-guided Crossmodal Entailment (PCE) which attempts to predict how a test person assesses the coherence of different modalities of a multimodal stimulus $x^{\text{\tiny STIM}}$ (e.g., an image with captions) given the test person's visual fixation sequence $x^{\text{\tiny ALIGN}}$. We collect the first POV learning benchmark data set based on PCE by tracing each test person's eye movements while this person is judging if the subjectively central visual objects are mentioned in the image's captions (see Sect.~\ref{Sec:PCE}).
    \item We empirically show how eye tracking sequences can be used for improved individual alignment of POV machine learning models. For this, we explore multiple ways to exploit the eye tracking data in a variety of machine learning models $f(X^{\text{\tiny POV}},Y^{\text{\tiny POV}})$ (see Sects.~\ref{sec:experiments} and \ref{sec:fixationModels}). To this end, we propose the Perception-Guided Multimodal Transformer (PGMT) which is optimized for PCE (see Sect.~\ref{sec:pgmt}).
    \item Furthermore, we assess how multimodal large language models (i.e.,~GPT4) perform on POV-learning in zero and few-shot in-context settings (see Sect.~\ref{subsec:gpt4}).
\end{enumerate}

Our findings suggest that augmenting machine learning models with individual perception signals and optimizing for POV-specific objectives is a promising direction for future research. 
We call this new perspective on AI alignment \emph{POV Learning}.

\section{Related work}
A variety of research areas are related to the work we present. On a technical level, the field of multi-modal (in our case visual-linguistic) representation learning provides the technical background for our experiments. The novel PCE task introduced in this work is related to the Visual Entailment tasks and adds a behavioral component to it.

\subsection{Large Language Models and their Limitations}
There are two key techniques on which many recent language models are based: the attention mechanism \cite{bahdanau2016neural} and the transformer architecture \cite{vaswani2017attention}. Despite their impressive capabilities, there are downsides associated with models like Gemini \cite{geminiteam2024gemini}, GPT4 \cite{openai2023gpt4}, Meta Llama 3
and beyond. Firstly, the amount of resources needed to train a modern language model in both time and storage is tremendous.
Secondly, the alignment of Large Language Models (LLMs) on an individual level as a POV learning task is an open research question.


\subsection{Visual-Linguistic Transformers and Tasks}
There exists a variety of recent approaches in Visual-linguistic representation learning. Before the introduction of novel multimodal LLMs like GPT4 \cite{openai2023gpt4} and Gemini \cite{geminiteam2024gemini}, which are currently the dominant approaches and outperform traditional models across all benchmarks, they usually combined the BERT architecture \cite{devlin2019bert} with a vision module. Models that fall in this category are VilBert \cite{lu2019vilbert}, VL-Bert \cite{su2020vlbert} and Visual Bert \cite{li2019visualbert}.

Others, like LXMERT \cite{tan2019lxmert}, Uniter \cite{chen2020uniter}, SimvLM \cite{wang2021simvlm}, and most recently, OFA \cite{wang2022ofa}, use other transformer-based architectures. 
They all have in common that they are evaluated at least partly on one or more of the benchmark tasks provided by datasets like Visual Genome \cite{krishnavisualgenome}, MS-Coco \cite{lin2015microsoft}, or Flickr30k \cite{plummer2016flickr30k}. Normally, the tasks are divided into two sub-categories which are briefly described here.
\begin{description}
    \item[Pre-training tasks:] Masked Language Modeling, Masked Region Modeling, Textual Grounding
    \item[Downstream tasks:] Visual Commonsense Reasoning, Visual Question Answering, Natural Language for Visual Reasoning, Region to Phrase Grounding, Visual Entailment
\end{description}
This list is not exhaustive, but contains tasks most commonly referred to in Visual-Linguistic representation learning. The task with the highest similarity to the Perception-guided Crossmodal Entailment task we propose in this work is Visual Entailment \cite{xie2019visual}. In Visual Entailment, an image text combination functions as a premise/hypothesis pair, and the task is to classify whether the hypothesis is entailed in the premise, contradicts the premise, or is neutral with regard to the premise.

\subsection{AI Alignment}
The goal of AI alignment is to ensure that AI models behave in a way that overlaps with human preferences. Due to the ever increasing impact of LLMs and AI models in general, behavior such as halluzinations and social biases must be mitigated to prevent harmful effects \cite{ji2024ai}.
To prevent such harmful responses, reinforcement learning from human feedback (RLHF) was originally introduced into the LLM deployment process. This technique relies on human-provided feedback, which serves as an alignment signal to prevent LLMs from returning harmful output \cite{ouyang2022training} and has since seen continuous improvements.

However, in this work, we divert from the global population-level view on alignment and instead focus on alignment on the individual level. Human values differ on all levels of granularity in human societies and cultures. The alignment on fine-grained levels down to the individual point-of-view has not been addressed so far. Here, we focus on an alignment signal that reflects individual behavior.

\subsection{Machine Learning on Eye Tracking Data}
Historically, eye tracking data have been most common as a signal in psychology of (visual) perception, where it has been used to determine how visual perception works, both physiologically and psychologically \cite{haber1980psychology}. 
In other lines of psychology-related work, eye tracking data are used to train statistical models to predict behavior in binary choice settings (cmp.~\cite{Krajbich2010VisualFA} \cite{10.3389/fnins.2017.00468} \cite{Bitzer2014PerceptualDM}). In contrast to our work, the works described above use directly derived information from the eye tracking sequences (i.e. dwell time, revisits) \cite{doi:10.1287/isre.2019.0907} as input for their statistical models, while we focus on symbolic representations of multimodal AOIs
that we use as input for deep learning models.

In summary, there are important key points in which our work differs from both the line of Visual-Linguistic Transformer research and the line of eye tracking based research in psychology. 
In contrast to Visual Entailment tasks, we focus on the behavioral aspect of visual perception which we use as individual alignment signal. In Visual Entailment, there is a single population-wide ground truth for a given image/text pair, whereas in the POV learning task, different participants might have opposing evaluations of a given image/text pair, which is why our models have to reflect this aspect in their inferences.
On the other side, the models in psychology do account for individual behavior, but are purely descriptive statistical models (as opposed to our inferential deep learning based approaches), use the data differently (feature engineering on the basis of some properties in the eye tracking sequence, vs. representation learning techniques on our end) and do not account for multi-modality.

Most importantly, no previous work attempted to fuse the two research areas in order to derive individually aligned machine learning models.

\section{Perception-guided Crossmodal Entailment}
\label{Sec:PCE}
This section describes how the data for both the perception-guided ML model exploiting eye tracking data $X^{\text{\tiny ALIGN}}$, as well as the content-based transformer models trained solely on the stimulus data $X^{\text{\tiny STIM}}$ were obtained. For both models, identical prediction targets $Y^{\text{\tiny POV}}$ are used, according to the task of Perception-guided Crossmodal Entailment (PCE).

As a basis for the eye tracking study, we selected a set of multimodal documents from the Visual Genome data set \cite{krishnavisualgenome}. Visual Genome (VG) consists of 108,249 images that are annotated on several layers, from regions to objects, represented in graph structures.
We leverage the underlying scene graph to rank entities depicted in an image by their centrality degree. Based on this ranking, we chose regions that are connected by their contained entities and retrieved the associated caption. 
The captions of the regions then serve as a textual description of the picture. The region descriptions combined with the image constitute the multi-modal stimulus data $X^{\text{\tiny STIM}}$.

\subsection{POV Task Specification}

The eye tracking study consisted of 153 manually selected stimuli. 
After presenting $X^{\text{\tiny STIM}}$ to the study participants, we asked them the question \textit{ ``Are the central objects in the image mentioned in the caption?''} resulting in their subjective evaluations $Y^{\text{\tiny POV}}$. We call this task \emph{Perception-guided Crossmodal Entailment} (PCE) since it requires cross-modal reasoning and assesses individual human perception.

\subsection{Eye tracking and Indvidual Assessment Recording}
\label{subsec:eye-tracking}
Overall, 109 participants in the age range of 19-61 years participated in the study, which took place in the eye tracking laboratory of the Media Studies department of our institution. Our institution's ethics board approved the experiment as no harm was to be expected for the participants when evaluating image captions. The majority of the participants were female (F= 77, 70.64 \%; M= 30, 27.52 \%), the average age was M\textsubscript{age} = 25.4 years. 
To ensure that the length of an eye tracking session was not too long to retain the participant's attention, the participants were divided into three groups.
A short instruction slide was shown followed by 50-53 blocks consisting of a text-image-stimulus and a following question slide. The stimulus and question slide blocks were shown in a randomized order. For the study, the experiment was programmed, recorded, and analyzed in the software iMotions (Version 9.1.5). The total duration for each session was approximately eight minutes (excluding the time for initial calibration).

\subsection{Symbolic Fixation Sequence Extraction}
\begin{figure}
    \begin{floatrow}
    \capbtabbox{%
    \small
      \begin{tabular}{l|llllll}
        \hline
        Nr. & AOI & Coordinates & Duration \\
        \hline
        \hline
        1 & $wall_{vis}$ & 865, 346 & 100.00 \\
        \hline
        2 & $wall_{text}$ & 709, 29 & 158.32 \\
        \hline
        ...&... & ... & \\
        \hline
        20 & $wall_{vis}$ & 988, 431 & 250.00 \\
        \hline
    \end{tabular}
    
    \caption{Fixation sequence for Participant EWCX and Stimulus ID 2412873.}
    \label{tableFixationData}
    }
    
    \ffigbox{
    \includegraphics[scale=0.25]{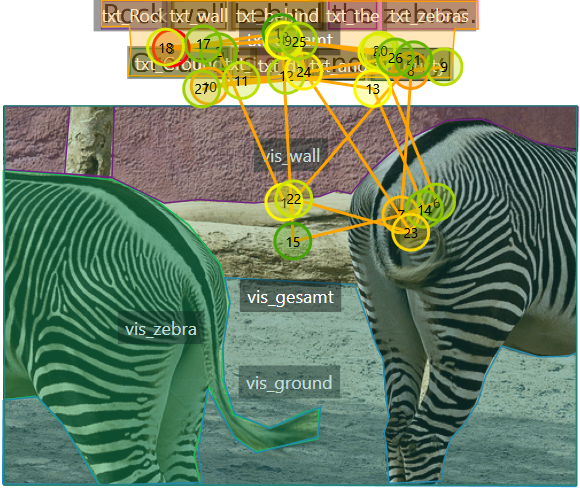}
    \caption{Fixation Sequence with a negative response.}%
    \label{fig2}
    }

    \end{floatrow}
\end{figure}

From the eye tracking study we obtain participant-specific fixation sequences $X^{\text{\tiny ALIGN}}$ for each stimulus $X^{\text{\tiny STIM}}$. In addition to the information about the participant and stimulus, each fixation contains information about its location, 
its duration, dispersion, and the associated annotated region. For example, Table \ref{tableFixationData} lists how the sequence in Figure \ref{fig2} is represented. All fixations are ordered by the time of their occurrence. 


\subsection{Data summary}
In summary, the data we obtain consist of the created stimuli (images, annotated AOIs and corresponding captions) $X^{\text{\tiny STIM}}$ a participant-specific fixation sequence $X^{\text{\tiny ALIGN}}$ and an individual PCE assessment $Y^{\text{\tiny POV}}$ for each stimulus. Overall, the acquired data set contains 5,400 unique fixation sequences, where each sequence represents a perception pattern of a participant generated when parsing the stimulus. The fixation sequences contain a total of 148,100 identified fixations, on average 27.42 fixations per stimulus exposure. The distribution of the prediction targets $Y^{\text{\tiny POV}}$ is heavily skewed, with 67.4 \% of the samples belonging to the \textit{Yes} class, 20.1 \% to the \textit{No} class, and 12.5 \% to the \textit{Unclear} class. The data set (including raw fixation data, stimuli, caption texts, and annotated regions) and the code for the experiments is available on GitHub.\footnote{https://github.com/siwer/Crossmodal-Entailment} 

\section{POV Models and Experimental Design}
\label{sec:experiments}
To test our hypothesis that the stimuli evaluation process $X^{\text{\tiny STIM}}$ is reflected in the eye tracking data $X^{\text{\tiny ALIGN}}$ and can act as a feature to improve individual alignment $g^{\text{\tiny POV}}$, three models $f()$ were implemented with focus on different aspects of $X$.

The focus of the \emph{LSTM} \cite{hochreiter1997long} model is on exploring the informational value of the fixation sequence of symbolic components of the multimodal stimulus as an alignment signal $x^{\text{\tiny ALIGN}_{\text{eye}}}$, without relying on content information $x^{\text{\tiny STIM}}$ on visual and textual entities in the stimuli. 
In contrast to this, the focus of the \emph{Transformer} model is on content-based information in the form of pre-trained feature representations of the stimuli $x^{\text{\tiny STIM}}$. In this way, we can quantify the influence of both components of $x^{\text{\tiny POV}}$ independently of each other.
To further test how individual preferences (without eye tracking information) impact performance, we also provide the participant ID as an alignment signal $x^{\text{\tiny ALIGN}_{\text{id}}}$ to the Transformer. 
Finally, as a combination of all, we train an \emph{Ensemble} model that evaluates the predictions of all models together to see if the signals complement each other. 

The task of all experimental setups
is to predict the evaluation of each participant of the text-image combinations individually. This problem can be formulated as a classification problem with three possible outcomes for $y^{\text{\tiny POV}}$:
\textit{1) \textbf{yes}, the caption mentions the central objects in the image}, \textit{2) \textbf{no}, central objects in the image are not mentioned} and \textit{3) \textbf{unclear}}. 

\subsection{Empirical Setup}
The data are divided into training, evaluation and test sets in a random 80/10/10 split, with each split retaining the global class distribution from the data.
All the models described in the following sections are optimized with AdamW \cite{loshchilov2019decoupled} optimizer on a cross-entropy loss and trained for a maximum of 30 epochs. 
The ranges of hyperparameters that were tested to find the best performing Ensemble and best performing Perception-guided Transformer are: LR (1e-04, 5e-04, 1e-05, 5e-05), FF-Dim (32, 64, 128, 256), Embedding Dim (8, 16, 32), Batch size (16,64,128). All possible combinations were evaluated. 
We report accuracy and F1 score as evaluation metrics $g^{\text{\tiny POV}}$ to account for class imbalance in the data. Due to the sparsity of the \textit{unclear} class, we tested all models in two different settings, one with all classes present (3 classes) and one with the \textit{unclear} class excluded from the evaluation.


\subsection{Perception-Guided LSTM Model}
\label{subsec:LSTM}
For this setup, we implement a unidirectional LSTM model that only receives symbolic representations of the AOIs as a sequence, called $x^{\text{\tiny ALIGN}_{\text{eye}}}$. In our case, we refer to all annotated regions of the eye tracking study as AOIs. If we take the sequence from Table \ref{tableFixationData} as an example, each row corresponds to one step in the LSTM. In this example, the symbol denoted \textit{vis\_wall} would correspond to the input in the first step t\textsubscript{1}, \textit{txt\_wall} to t\textsubscript{2}, and so on until the last element of the sequence \textit{vis\_wall} is reached.
Before feeding the sequence into the LSTM, we put the one-hot encoded elements through an embedding layer that outputs a dense low-dimensional vector representation of the elements in the AOI sequence. Optionally, we also apply the same technique to create embeddings for the participants, called $x^{\text{\tiny ALIGN}_{\text{id}}}$. As a final step, we concatenate each AOI sequence element with the current participant embedding and put it through a feedforward layer to receive a combined representation, capturing \textit{Who is looking at which AOI}. This final representation is then used as input for the LSTM, of which we take the final hidden state and perform the classification $\hat{y}^{\text{\tiny POV}}$ by running it through a feedforward layer.
During training, the embedding layers are also updated, so that the resulting representations are fitted to the task. Overall, there are 837 unique AOI identifiers (i.e., the text representations like "vis\_wall" or "txt\_zebra") in the dataset. 
This setting can be seen as using only the individual alignment signal $x^{\text{\tiny ALIGN}}$, without the stimulus content information beyond its symbolic abstraction $x^{\text{\tiny STIM}}$.

\subsection{Content-based Transformer Model}
\label{subsec:transformer}
\begin{figure}
    \begin{floatrow}
    \capbtabbox{%
    \small
    \begin{tabular}{l|l|l|l|l}
        & \textbf{wall\textsubscript{\tiny vis}} & \textbf{wall\textsubscript{\tiny txt}} & \textbf{off} & \textbf{rock\textsubscript{\tiny txt} \tiny ...} \\
        \hline
       \textbf{wall\textsubscript{\tiny vis}} & 0 & 1 & 0 & 0 \tiny ...\\
        \hline
        \textbf{wall\textsubscript{\tiny txt}} & 0 & 0 & 1 & 0 \tiny ...\\
        \hline
        \textbf{off} & 0 & 0 & 0 & 1 \tiny ...\\
        \hline
        \textbf{rock\textsubscript{\tiny txt}} & 0 & 1 & 0 & 0 \tiny ...\\
        \hline
        \textbf{\tiny ...} & \tiny ... & \tiny ... & \tiny ... & \tiny ... 
    \end{tabular}
    
    \caption{Transition Matrix (partial) for Participant EWCX and Stimulus ID 2412873.}
    \label{seqToMatrix}
    }
    
    \ffigbox{
    \includegraphics[scale=0.25]{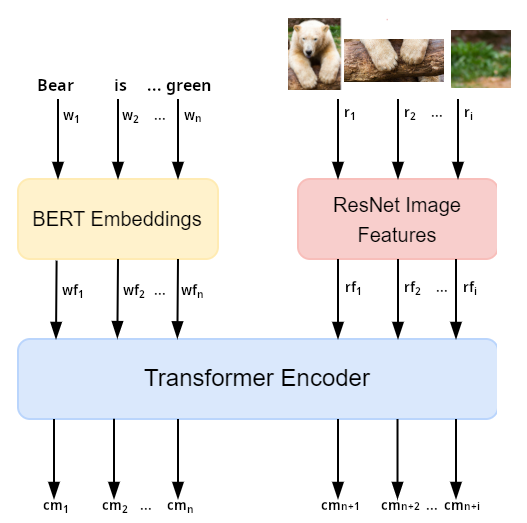}
    \caption{Baseline Multimodal Transformer: Textual input is referred to as $w$ and image regions as $r$, respectively. The outputs marked as $cm$ are cross-modal embeddings carry both visual and linguistic features}%
    \label{fig:fig1}
    }

    \end{floatrow}
\end{figure}

The fundamental building block for this baseline is a transformer encoder, which comprises multiple attention heads and layers. Figure \ref{fig:fig1} shows the baseline model, which can be extended and adapted to different tasks by adding additional layers on top. 
We use pre-trained BERT\footnote{https://huggingface.co/docs/transformers/v4.28.1/en/model\_doc/bert} \cite{devlin2019bert} embeddings that are obtained from the \textit{Huggingface Transformers} Package \cite{wolf2020huggingfaces} as representations for the linguistic inputs (the captions) and a pre-trained ResNet50\footnote{https://pytorch.org/vision/main/models/resnet.html} \cite{he2015deep} to obtain image features of the predefined regions. 

For the experiments described in the following, we additionally concatenated a low-dimensional participant representation to the cross-modal classification token, as a rudimentary alignment signal. There is a single feed-forward layer between the ResNet image features and the transformer encoder, which is used to downsize the image feature dimension to match the BERT Embedding dimension. For the final decision $\hat{y}^{\text{\tiny POV}}$, the classification token is put through a single feed-forward layer with three output dimensions that correspond to the three possible outcomes. Note that the ResNet and BERT weights are not updated during training.
Conceptually, the main advantage of the Transformer architecture is that it can use background knowledge in form of the pre-trained representations of the input modalities to solve the Perception-guided Crossmodal Entailment task.

\subsection{Ensemble Model}
In addition to the perception-guided approach of the LSTM and the content-based approach of the Transformer model, we introduce a meta-model as a combination of both. In this ensemble model, we simply combine both estimates of the models and use the average as our final classification $\hat{y}^{\text{\tiny POV}}$. 

\section{Experiments I - The Benefit of Alignment Signals in Basic POV Models}
\label{sec:fixationModels}
In a first set of experiments, we investigated how the different types of information (content $x^{\text{\tiny STIM}}$  vs. perception $x^{\text{\tiny ALIGN}})$ contribute to solving the PCE task when using the basic ML models introduced in Sect.~\ref{subsec:LSTM} and Sect.~\ref{subsec:transformer}. 
For a fair comparison, all three approaches are trained using the same hyperparameters, based on the best-performing ensemble model. 

We use 6 attention heads and layers for the transformer, 256 dimensions for the transformer feed-forward network and 32 embedding dimensions (for both the participant and AOI embedding) in the LSTM network. All three models are trained using a batch size of 128 and a learning rate of 0.0001. We report the prediction accuracy and the F1 score (averaged over all classes) on the test split as evaluation metrics $g^{\text{\tiny POV}}$ for all conducted experiments. Due to the rarity of the \textit{unclear} class, we also report accuracy and F1 scores that exclude samples with this class ('2 Class' setting). In addition to our models, we report a naive baseline that only predicts the most frequent class. Table \ref{table:overview} shows the effect of different combinations of sequential and contextual information and functions as an ablation. It can be cleary seen that a combination of all available signals lead to the best results.



These results clearly confirm our hypothesis that the fixation sequence contains information that can be used as an alignment signal $x^{\text{\tiny ALIGN}_{\text{eye}}}$ in a POV learning task. The naive baseline only reached half of the F1 score of the worst model. In addition, they show that the combination of $x^{\text{\tiny STIM}}$ and sequential information in an ensemble model leads to the best results, indicating that they complement each other.

To test the importance of providing a participant id $x^{\text{\tiny ALIGN}_{\text{id}}}$ to enable the personalization of the alignment signal, we investigate how much the individual participant embeddings (see Sect.~\ref{subsec:LSTM}) of our models contribute to the overall performance. 
It becomes apparent that even the simple representation of the participants that we learn within our models provides a useful alignment signal $x^{\text{\tiny ALIGN}_{\text{id}}}$ and improves the individual alignment.


\begin{table}
    \caption{Effects of the presence of alignment signals}\label{table:overview}
    \centering
    \small
    \begin{tabular}{l|l|l|l|l|l|l|l}
    Model & Eyetrack & User & Stimulus & \multicolumn{2}{|c|}{3 classes} & \multicolumn{2}{|c}{2 classes}\\
    $f()$ & $x^{\text{\tiny ALIGN}_{\text{eye}}}$ & $x^{\text{\tiny ALIGN}_{\text{id}}}$ & $x^{\text{\tiny STIM}}$ & F1 & Acc. & F1 & Acc.\\
    \hline
    \hline
    Naive & & & & .2678 & .6715 & .4330 & .7639\\
    \hline
    LSTM & + &  &  & .3849 & .6944 & .6144& .7899\\
    \hline
    LSTM & + & + &  & .4849 & .7466 & .7767& .8494\\
    \hline
    Transformer &  &  & + & .5146 & .7189 & .7742& .8254\\
    \hline
    Transformer &  & + & + & .5240 & .7238 & .7879 & .8611\\
    \hline
    Ensemble & + & + & + & \textbf{.5628} & \textbf{.7728} & \textbf{.8112}& \textbf{.8742}\\
    \hline
    \end{tabular}
\end{table}

\section{Experiments II - The Perception-Guided Multimodal Transformer}
\label{sec:pgmt}
We propose a more efficient approach to combining content and sequential information by directly adding the fixation sequences in the form of transition matrices to the attention weights in the multimodal Transformer model. The basic idea behind this is to guide the attention heads to concentrate on AOI pairs that appear consecutively in the perception traces.
We call this a Perception-Guided Multimodal Transformer (PGMT). An additional advantage of this procedure is that it requires a reduced number of trainable parameters when compared to the vanilla transformer approach and no additional model as the LSTM in the ensemble approach. Table \ref{seqToMatrix} shows the transition matrix derived from the first 5 rows of table \ref{tableFixationData}.


We use the \textit{src\_mask} parameter in the PyTorch Transformer Encoder to add the transition matrix to the attention weights.
Due to the sparsity of the matrices and the final softmax application that happens after adding the matrices to the weights, it is beneficial to scale the weights to ensure that the perception information is not vanishing during calculations. This is enabled by the manual tuning parameter $\lambda$. We also found it beneficial to transpose the matrix along the diagonal and add the result to the original matrix. 
 $$   \text{Amplify}(M\textsubscript{Transition},\lambda)=\lambda(M\textsubscript{Transition} + M\textsubscript{Transition}^T) 
 $$
An example based on the matrix in Table \ref{seqToMatrix} is shown below:
When we input the matrix from table \ref{seqToMatrix} 
and set the weight variable \textit{$\lambda$} to 5, we get:

$$    \text{Amplify}(M\textsubscript{Transition},5)
    =
    \left[\begin{array}{rrrr} 
    0 & 5 & 0 & 0\\ 
    5 & 0 & 5 & 5\\ 
    0 & 5 & 0 & 5\\
    0 & 5 & 5 & 0\\
    \end{array}\right]
$$
This operation reduces the sparsity of the matrix and in combination with upscaling leads to the best results.

In this setup, we used 128 feed-forward dimensions in the transformer, a batch size of 16 and a learning rate of 0.0001. The transition matrices are weighted by \textit{$\lambda$} $= 500$, which was manually tuned and was most effective in this scenario.
When evaluated on the test data, the PGMT achieves an on-par performance with the Ensemble model in both settings. Table \ref{PGT-results} shows that the approach leads to competitive results, while requiring less trainable parameters and no separate model to capture sequential information as in the ensemble.


\section{Experiments III - In-Context POV Learning with LLMs}
\label{subsec:gpt4}
So far, we have shown that different ML models can be efficiently trained to benefit from POV signals $x^{\text{\tiny ALIGN}}$ resulting in improved predictive performance. Large Language Models (LLMs) have the potential to generate predictions, without training their parameters, just by prompting them with demonstrations of $x^{\text{\tiny ALIGN}}$.

With two state-of-the-art multi-modal large language models, we investigate how well they perform on POV tasks like PCE and if they are able to utilize perception information to improve individual alignment through few-shot learning (i.e., without any form of post-training).
We investigate i) how well multimodal large language models like GPT4 and Pixtral are capable of solving our PCE task, and ii) if few-shot in-context learning can improve their performance. As before, the task for the model is to predict how the individual participants evaluated the question \textit{'Does the caption mention the central entities in the image?'}, given a certain stimulus. As in the experiments before, there are three possible outcomes: \textit{Yes}, \textit{No}, \textit{Unclear}. 
The different setups vary in the amount of participant-specific information $X^{\text{\tiny ALIGN}}$ provided in the prompt. The data and the training and test splits are the same as before: 
\begin{description}
    \item [Zero-shot:] No participant-specific information $x^{\text{\tiny ALIGN}}$ is provided. The model is tasked with evaluating the question based only on the stimulus $x^{\text{\tiny STIM}}$, completely ignoring individual evaluations of the stimuli.
    \item [Fixations:] In addition to the stimulus itself, the subject-specific fixation sequence for the current stimulus $x^{\text{\tiny ALIGN}}$ is provided in the prompt.
    \item [One-shot:] This setup provides an additional training sample $x^{\text{\tiny ALIGN}}$ as a demonstration of the current individual's behavior and adds it to the system prompt. 
    Thus, each prompt additionally contains a stimulus, a corresponding fixation sequence, and the individual response.
\end{description}

As before, we report Accuracy and the F1 score in both the 2 class 
and the 3 class setting as performance metrics for the experiments. The experiment was carried out with requests to the OpenAI API and with a local instance of Pixtral\footnote{https://huggingface.co/mistralai/Pixtral-12B-2409}.


\begin{figure}
    \begin{floatrow}
    \capbtabbox{%
    \small
    \begin{tabular}{ll|l}
        Setting & Metric & Result \\
        \hline
        \hline
        Ensemble &Acc. &  \textbf{.8742}\\
        (2 Class)&   F1 &  .8112\\
        \hline
        PGMT & Acc.& .8715\\
        (2 Class)& F1 & \textbf{.8196}\\
        \hline
        \hline
        Ensemble & Acc. &  .7728 \\
        (3 Class)&  F1  & \textbf{.5628}  \\
        \hline
        PGMT &  Acc.& \textbf{.7761} \\
        (3 Class)& F1 & .5294 \\
        
    \end{tabular}
    
    \caption{Comparison of the Perception-Guided Multimodal Transformer with the Ensemble model.}
    \label{PGT-results}
    }
    
    \capbtabbox{%
    \small
    \begin{tabular}{ll|llll}
        Setting & Metric & Zero-shot & Fixation & One-shot \\
        \hline
        \hline
        Pixtral & Acc. & .7340 & .7439 & .5180 \\
        (2 Class) & F1 & .5208 & .4985 & .4485 \\
        \hline
        GPT4 & Acc. & .7472 & \textbf{.7657} & .7490 \\
        (2 Class) & F1 & .5156 & \textbf{.5225} & .5018 \\
        \hline
        \hline
        Pixtral & Acc. & .6454 & .6535 & .4460 \\
        (3 Class) & F1 & .3251 & .3110 & .2770 \\
        \hline
        GPT4 & Acc. & .6568 & \textbf{.6732} & .6584 \\
        (3 Class) & F1 & .3220 & \textbf{.3265} & .3132 \\
    \end{tabular}
    
    \caption{Experimental Results for LMM alignment with GPT-4 and Pixtral 12B.}
    \label{table:gpt4Alignment}
    }

    \end{floatrow}
\end{figure}


There are two noteworthy outcomes regarding our LMM research questions: Concerning ii) it appears that the LLMs cannot benefit from few-shot learning for the PCE task. At least not in the way we provided fixation information and demonstrations of individual behavior $X^{\text{\tiny ALIGN}}$ in the prompt. A perception trace and the PCE task might be too different and complex to tasks commonly used in post-training. However, in-context learning with the fixation sequence only improves performance. Concerning i) we found that overall LLM performance in PCE is considerably and consistently lower than all approaches presented before. This suggests that the tested LLMs have not been instruction-tuned on a task similar to PCE and that they struggle with this task in general and with individual alignment in particular. Obviously, more research is needed to assess if post-training helps for POV tasks.

\section{Discussion}
In this paper, we introduce and formalize \emph{POV Learning}. By demonstrating the potential of taking a subject's point-of-view and incorporating individual characteristics into machine learning models, we can improve predictive performance on an individual level. We also showed that state-of-the-art multimodal models like GPT4 struggle with alignment on an individual level on our newly introduced Perception-guided Crossmodal Entailment (PCE) task. 

As opposed to alignment techniques like RLHF \cite{ouyang2022training}, which tackles alignment as a population-level problem, our work interprets alignment on an individual level.

In three experiments, we explored different ways of using human perception sequences and symbolic representations based on user IDs as alignment signals for POV-learning tasks.  
To determine the relevance of the information contained in a fixation sequence, we implemented an LSTM model that processes a purely symbolic representation of the AOIs in the order of the fixations. This information alone leads to solid results in the PCE task, with an accuracy value of \textbf{.7466}, compared to \textbf{.6715} for a naive model and \textbf{.7238} for a content-only transformer model. Combining the LSTM with a Transformer that processes additional contextual information in the form of pre-trained features of the text and image batches of the multimodal stimulus, the resulting Ensemble model achieves a further improvement up to an accuracy of \textbf{.7728}. From this experiment, we conclude that the usage of human perception data leads to improved individual alignment in the PCE task. 

Additionally, we propose the Perception-Guided Multimodal Transformer (PGMT), which manipulates the attention weights according to sequence information, which improves Accuracy to \textbf{.7761}. In comparison to the Ensemble model, this approach has the advantage that the extra parameters of the LSTM model can be avoided while the number of parameters in the Transformer stays the same. With a maximum Accuracy of \textbf{.6732}, the LLMs are clearly outperformed by the PGMT and struggle with both the PCE tasks and in-context learning in this setting.


In summary, we provided a new benchmark data set for the novel Perception-guided Crossmodal Entailment task PCE for POV Learning and proposed the PGMT model, which can be used to explore the possibilities of aligning human-like perception with transformer models by exploiting individual alignment signals.

\section{Future work}
The most obvious direction for future work in POV learning is towards the generalization of the presented approaches to other forms of individual alignment signals. In a first step, a reliable and less costly method (in comparison to conducting an eye tracking study) for capturing individual human alignment has to be investigated. Ultimately, the benefits of perception-guided POV models must be tested in settings where multimodal stimuli are observed ``in the wild''. Identifying and obtaining valuable alignment signals is a future challenge for further research in the field of individual alignment and POV learning.

Another promising future study could be concerned with improving the performance of Multimodal LLMs like GPT4. So far, the proposed task and the few-shot prompts did not result in a convincing performance. Additional approaches to aligning MLLMs' inferences to individual preferences via post-training need to be investigated.

\bibliographystyle{vancouver}
\bibliography{hhai25}

\begin{thebibliography}{10}

\bibitem{openai2023gpt4}
OpenAI. GPT-4 Technical Report; 2023.

\bibitem{agrawal2024pixtral12b}
Agrawal P, Antoniak S, Hanna EB, Bout B, Chaplot D, Chudnovsky J, et~al.. Pixtral 12B; 2024.

\bibitem{bahdanau2016neural}
Bahdanau D, Cho K, Bengio Y. Neural Machine Translation by Jointly Learning to Align and Translate; 2016.

\bibitem{vaswani2017attention}
Vaswani A, Shazeer N, Parmar N, Uszkoreit J, Jones L, Gomez AN, et~al.. Attention Is All You Need; 2017.

\bibitem{geminiteam2024gemini}
Team G. Gemini: A Family of Highly Capable Multimodal Models; 2024.

\bibitem{devlin2019bert}
Devlin J, Chang MW, Lee K, Toutanova K. BERT: Pre-training of Deep Bidirectional Transformers for Language Understanding; 2019.

\bibitem{lu2019vilbert}
Lu J, Batra D, Parikh D, Lee S. ViLBERT: Pretraining Task-Agnostic Visiolinguistic Representations for Vision-and-Language Tasks; 2019.

\bibitem{su2020vlbert}
Su W, Zhu X, Cao Y, Li B, Lu L, Wei F, et~al.. VL-BERT: Pre-training of Generic Visual-Linguistic Representations; 2020.

\bibitem{li2019visualbert}
Li LH, Yatskar M, Yin D, Hsieh CJ, Chang KW. VisualBERT: A Simple and Performant Baseline for Vision and Language; 2019.

\bibitem{tan2019lxmert}
Tan H, Bansal M. LXMERT: Learning Cross-Modality Encoder Representations from Transformers; 2019.

\bibitem{chen2020uniter}
Chen YC, Li L, Yu L, Kholy AE, Ahmed F, Gan Z, et~al.. UNITER: UNiversal Image-TExt Representation Learning; 2020.

\bibitem{wang2021simvlm}
Wang Z, Yu J, Yu AW, Dai Z, Tsvetkov Y, Cao Y. SimVLM: Simple Visual Language Model Pretraining with Weak Supervision; 2021.

\bibitem{wang2022ofa}
Wang P, Yang A, Men R, Lin J, Bai S, Li Z, et~al.. OFA: Unifying Architectures, Tasks, and Modalities Through a Simple Sequence-to-Sequence Learning Framework; 2022.

\bibitem{krishnavisualgenome}
Krishna R, Zhu Y, Groth O, Johnson J, Hata K, Kravitz J, et~al.. Visual Genome: Connecting Language and Vision Using Crowdsourced Dense Image Annotations; 2016.

\bibitem{lin2015microsoft}
Lin TY, Maire M, Belongie S, Bourdev L, Girshick R, Hays J, et~al.. Microsoft COCO: Common Objects in Context; 2015.

\bibitem{plummer2016flickr30k}
Plummer BA, Wang L, Cervantes CM, Caicedo JC, Hockenmaier J, Lazebnik S. Flickr30k Entities: Collecting Region-to-Phrase Correspondences for Richer Image-to-Sentence Models; 2016.

\bibitem{xie2019visual}
Xie N, Lai F, Doran D, Kadav A. Visual Entailment: A Novel Task for Fine-Grained Image Understanding; 2019.

\bibitem{ji2024ai}
Ji J, Qiu T, Chen B, Zhang B, Lou H, Wang K, et~al.. AI Alignment: A Comprehensive Survey; 2024.

\bibitem{ouyang2022training}
Ouyang L, Wu J, Jiang X, Almeida D, Wainwright CL, Mishkin P, et~al.. Training language models to follow instructions with human feedback; 2022.

\bibitem{haber1980psychology}
Haber RN, Hershenson M.
\newblock The Psychology of Visual Perception.
\newblock Holt, Rinehart and Winston; 1980.

\bibitem{Krajbich2010VisualFA}
Krajbich I, Armel C, Rangel A.
\newblock Visual fixations and the computation and comparison of value in simple choice.
\newblock Nature neuroscience. 2010;13 10:1292-8.

\bibitem{10.3389/fnins.2017.00468}
Tavares G, Perona P, Rangel A.
\newblock The Attentional Drift Diffusion Model of Simple Perceptual Decision-Making.
\newblock Frontiers in Neuroscience. 2017;11.

\bibitem{Bitzer2014PerceptualDM}
Bitzer S, Park H, Blankenburg F, Kiebel SJ.
\newblock Perceptual decision making: drift-diffusion model is equivalent to a Bayesian model.
\newblock Frontiers in Human Neuroscience. 2014;8.

\bibitem{doi:10.1287/isre.2019.0907}
Pfeiffer J, Pfeiffer T, Mei\ss{}ner M, Wei\ss{} E.
\newblock Eye-Tracking-Based Classification of Information Search Behavior Using Machine Learning: Evidence from Experiments in Physical Shops and Virtual Reality Shopping Environments.
\newblock Information Systems Research. 2020;31(3):675-91.

\bibitem{hochreiter1997long}
Hochreiter S, Schmidhuber J.
\newblock Long short-term memory.
\newblock Neural computation. 1997;9(8):1735-80.

\bibitem{loshchilov2019decoupled}
Loshchilov I, Hutter F. Decoupled Weight Decay Regularization; 2019.

\bibitem{wolf2020huggingfaces}
Wolf T, Debut L, Sanh V, Chaumond J, Delangue C, Moi A, et~al.. HuggingFace's Transformers: State-of-the-art Natural Language Processing; 2020.

\bibitem{he2015deep}
He K, Zhang X, Ren S, Sun J. Deep Residual Learning for Image Recognition; 2015.

\end{thebibliography}
\end{document}